\documentclass[10pt,twocolumn,letterpaper]{article}

\usepackage{wacv}              

\usepackage{tabularx}

\usepackage{algorithm}
\usepackage{algpseudocode}
\usepackage{enumitem}
\algnewcommand{\INPUT}{\item[\textbf{Input:}]}
\algnewcommand{\OUTPUT}{\item[\algorithmicoutput]}
\algnewcommand{\algorithmicoutput}{\textbf{Output:}}

\usepackage{amsmath}

\usepackage{graphicx}
\usepackage{amsmath}
\usepackage{amssymb}
\usepackage{booktabs}
\usepackage{subcaption}
\usepackage{float}
\usepackage{verbatim}
\usepackage[pagebackref,breaklinks,colorlinks]{hyperref}

\DeclareMathOperator{\PCEEM}{PCEEM}
\DeclareMathOperator{\LRP}{LRP}
\usepackage[capitalize]{cleveref}
\crefname{section}{Sec.}{Secs.}
\Crefname{section}{Section}{Sections}
\Crefname{table}{Table}{Tables}
\crefname{table}{Tab.}{Tabs.}

\def\wacvPaperID{1983} 
\def\confName{WACV}
\def\confYear{2025}

\begin{document}

\title{Efficient Progressive Image Compression with Variance-aware Masking}

\author{
\begin{tabular}{c}
Alberto Presta$^{1}$ \quad Enzo Tartaglione$^{2}$ \quad Attilio Fiandrotti$^{1,2}$ \quad Marco Grangetto$^{1}$ \quad Pamela Cosman$^{3}$
\end{tabular}
\\
$^1$University of Turin, Italy  ~~ \\ $^2$ LTCI, T\'el\'ecom Paris, Institut Polytechnique de Paris~~ \\
$^3$ Dept. of Electrical and Computer Engineering, UC San Diego, CA, USA
\\
 {\tt\small alberto.presta@unito.it}
}

\maketitle

\begin{abstract}

Learned progressive image compression is gaining momentum as it allows improved image reconstruction as more bits are decoded at the receiver.
We propose a progressive image compression method in which an image is first represented as a pair of base-quality and top-quality latent representations.
Next, a residual latent representation is encoded as the element-wise difference between the top and base representations.
Our scheme enables progressive image compression with element-wise granularity by introducing a masking system that ranks each element of the residual latent representation from most to least important, dividing it into complementary components, which can be transmitted separately to the decoder in order to obtain different reconstruction quality. 
The masking system does not add further parameters or complexity.
At the receiver, any elements of the top latent representation excluded from the transmitted components can be independently replaced with the mean predicted by the hyperprior architecture, ensuring reliable reconstructions at any intermediate quality level.
We also introduced Rate Enhancement Modules (REMs), which refine the estimation of entropy parameters using already decoded components.
We obtain results competitive with state-of-the-art competitors, while significantly reducing computational complexity, decoding time, and number of parameters.

\end{abstract}

\section{Introduction} \label{sec:intro}

In recent years, Learned Image Compression (LIC) has attracted significant interest, outperforming standardized codecs~\cite{jpeg,jpeg2k,vvc} in Rate-Distortion (RD) efficiency for natural images~\cite{zou2022devil,liu2023tcm}.
In a learnable codec, an encoder on the transmitter side projects an image to a latent space that is quantized and entropy coded into a compressed bitstream.
On the receiver side, the bitstream is processed by a decoder that reverses the encoding process, recovering (a distorted version of) the original image.

\begin{figure}[t!]
  \centering
 \includegraphics[width=0.97\columnwidth]{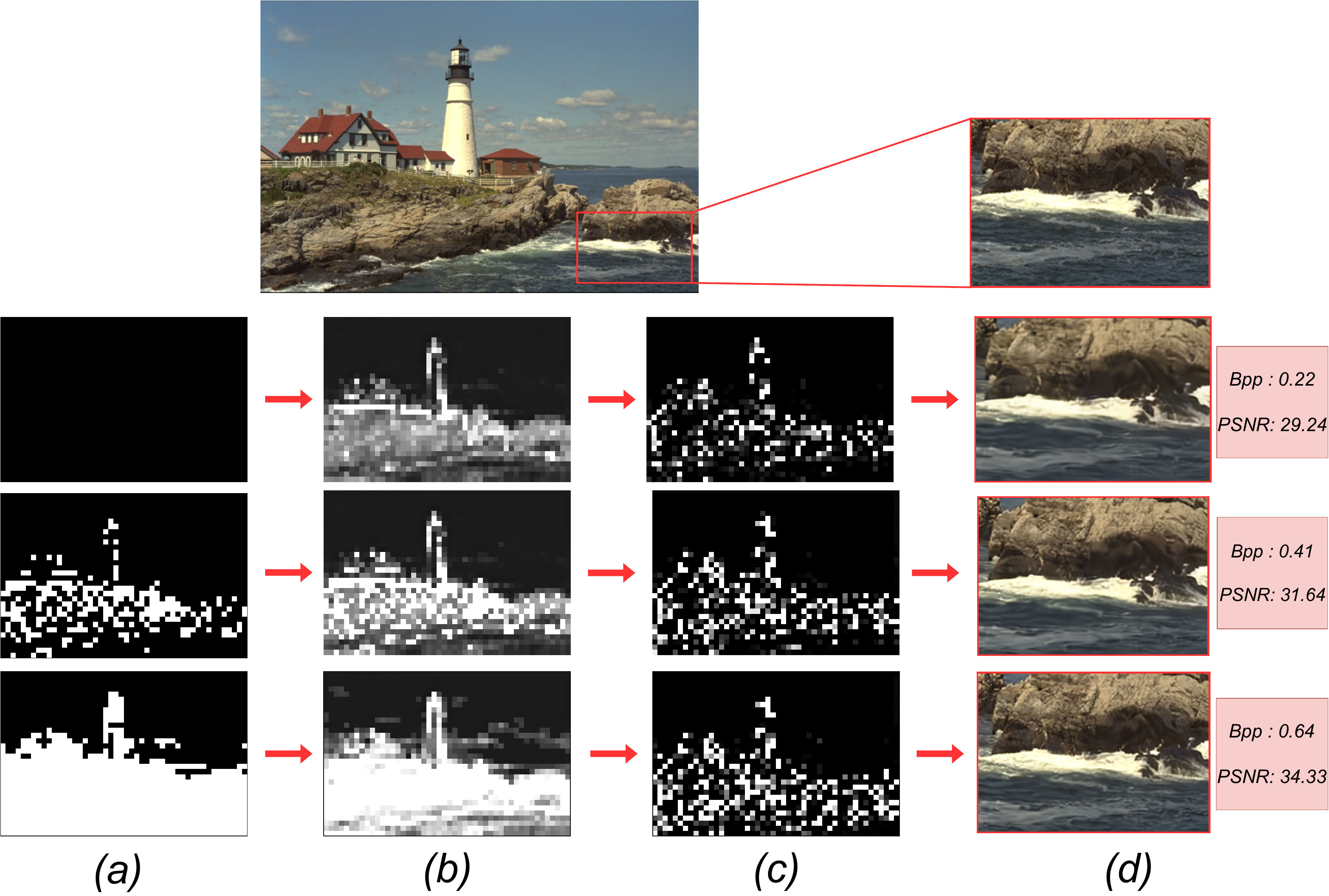}
 \vspace{-0.3cm}
  \caption{Compression results for three different qualities, which increase across rows.
  Adding details via the masking system (a) increases the standard deviation in the non-masked latent representation (b)  to add  details  (c) for a better reconstruction (d).  }
  \label{fig:teaser}
  \end{figure}

However, LIC codecs still face the challenge of meeting the rate of channels whose capacity changes as a function of connection type and congestion at the nodes.
Scalable coding \cite{ohm2005advances} consists in encoding a content as one base bitstream enabling to recover a low-quality version of the content and a few enhancement bitstreams enabling improved quality when received.
With progressive image compression,  sometime known also as fine-grained scalability (FGS), this concept is further extended improving the quality of the reconstructed image as each extra bit from the same bitstream is received \cite{li2001overview}, allowing to truncate it at.
A few learnable image compression schemes with progressive decoding properties have been known to exist to date; some allow adjusting the tradeoff between compressed bitrate and quality by exploiting a single rate-variable model~\cite{cui2021asymmetric,sel_comp, kamisli2024variable}. However, a different bitstream must be encoded and delivered for each different bitrate target.
Early models such as Diao \emph{et al.} \cite{diao2020drasic} and Lu \emph{et al.} \cite{lu2021progressive} achieved progressive representation through distributed recurrent autoencoder and nested quantization, respectively.
Lee~\emph{et~al.}~\cite{lee2022dpict} introduced a novel way to represent each element of the latent representation based on trit-plane coding, achieving state-of-the-art RD efficiency, without exploiting any kind of context model, which were introduced by~\cite{jeon2023context} for both rate and distortion reduction. 
A RD prioritized transmission system was introduced to be able to find the trit-planes with more information and give them priority for being encoded first.
However, exploiting such modules for both encoding and decoding and finding the right priority transmission distribution make \cite{jeon2023context} and \cite{lee2022dpict} costly in terms of computational resources and time. 

In this work, we propose a learnable, efficient, and progressive image compression architecture.
It has two initial levels, \emph{base} and \emph{top}; the first defines the lowest quality while the second allows the system to achieve all the other higher qualities.
To achieve FGS, we first compute a residual latent representation by means of element-wise difference between the top and base ones, and then break it down into complementary parts, which form the final bitstream, which can be encoded and sent separately, resulting in reconstruction at multiple qualities, as shown in Fig. \ref{fig:teaser}. 
Furthermore, we added in the main architecture learnable rate enhancement modules (REMs) to further improve the estimation of entropy parameters.
Our progressive approach is built on a channel-wise entropy parameter module \cite{minnen2020channel}. 

The main contribution of this paper are the follows:
\vspace{-0.1cm}
\begin{itemize}[itemsep=0.01em] 
    \item We introduced a method where, to achieve FGS, complementary portions of a residual latent representation are added to the \emph{base} one, which represents the lowest level in terms of bitrate. 
    \item We exploited a lightweight masking policy which ranks the elements of the residual latent representation from most to least important, also identifying their positions, which allows for the creation of a progressive system where more elements are added incrementally.
    \item  We achieved competitive results with respect to~\cite{jeon2023context} in RD performance, significantly reducing complexity in terms of computational resources, decoding time, and number of parameters.
 
\end{itemize}

\section{Related work}
\subsection{Learning-based image compression}
While seminal works about learning-based image compression are based on recurrent neural networks~\cite{gregor2016towards,toderici2015variable,toderici2017full}, those based on variational autoencoders (VAEs)~\cite{kingma2013auto} have gained significant importance.
Initially, the general architecture is formed by a simple autoencoder~\cite{balle2016end, theis2017lossy} where the latent representation is modeled with a channel-wise probability distribution, extracted analytically~\cite{stf23} or with a neural network.
Quantization is replaced by uniform additive noise during training; this effective technique became the most common practice in such models. 
To further improve the entropy estimation of the latent elements, a hyperprior autoencoder is added to the main architecture~\cite{balle2018variational, minnen2018joint}, with the aim of finding spatial correlation within the image. In such a scenario, the main latent representation is modeled with a Gaussian distribution. More effort has been put into optimizing the entropy estimation by introducing a parallelizable checkerboard context model~\cite{He_2021_CVPR} or a channel-wise approach~\cite{minnen2020channel}. The latter divides the latent space into blocks, called slices, along the channel dimension. For example, if the latent space has 320 channels and 10 slices are desired, each slice contains 32 channels. These slices are sequentially encoded, exploiting previous ones; this method can be seen as an autoregressive model along channels, improving efficiency of such models, and enhancing entropy parameter estimation.
Other alternative techniques have been used, such as invertible modules~\cite{xie2021enhanced}, sparse representation spanned by learned visual codebooks~\cite{jiang2024neural}, or graph-based attention~\cite{spadaro2024gabic}.

Other works tried to exploit an attention mechanism to optimize the bits allocation, by integrating non-local attention blocks in the image compression architecture~\cite{cheng2020learned, zhou2019end, liu2019non}. 
In~\cite{zou2022devil}, the authors replaced the non-local attention block introduced in~\cite{liu2019non} with a window block computed from spatially neighboring elements that focus on image details, proving that this can improve RD performance with less computation.
Similarly, the model in~\cite{liu2023tcm} combines Transformer and CNN blocks to leverage both local and non-local modeling abilities. They also improved the entropy estimation module by adding a Swin-attention module.
\subsection{Progressive image compression}

In progressive image compression, reconstruction quality improves
as more bits of the bitstream are received and decoded.
To achieve this property, early works exploited recurrent neural networks (RNNs)~\cite{medsker2001recurrent}.
The model in~\cite{toderici2015variable} exploits long-short memory~\cite{hochreiter1997long} to transmit bits progressively. The encoder and decoder structures were improved in~\cite{toderici2017full} for higher resolution images. In~\cite{gregor2016towards} a recurrent model is exploited, enhancing perceptual quality through a generative model; however, this method is suitable only for low resolution patches.
In \cite{diao2020drasic}, a distributed recurrent auto-encoder was designed for scalable image compression, while
\cite{johnston2018improved} introduced an iterative process to generate the binary codec and supported spatially adaptive bit rates that dynamically adjust bit allocation based on context.

In \cite{cai2019novel}, a new architecture with a single encoder and two decoders was introduced. This setup produces two representations: a preview image from the first decoder and a full-quality image from the second one. Although it uses two representations like our model, \cite{cai2019novel}~relies on concatenation rather than residual addition, thus doubling the input dimension of the second decoder.
Furthermore, they did not introduce a technique for sending partial portions of the latent space. Due to these limitations, \cite{cai2019novel}~supports only two levels of quality.
These methods not only have worse performance than standard codecs, but also support only coarse granular scalability.
In \cite{wang2023evc}, several progressive encoders were introduced, one for each target quality, to achieve different levels of reconstruction; however, they do not provide FGS, and this method requires training separate encoders for each desired target quality.

To obtain FGS, \cite{lu2021progressive}~introduced a nested quantization, obtaining results comparable to conventional deep image codecs.
\cite{lee2022dpict}~proposed a novel trit-plane coding method, which represents each element of the latent space with a ternary number. Specifically, they use $L$ trits to express each element, ordered from the most significant to the least significant trit. In that sense, they also introduced an RD-prioritized transmission system that optimizes the order of trit transmission based on their priorities. 
This method was improved in \cite{jeon2023context} by introducing a context model capable of improving both the estimation of the entropy parameters and the quality of reconstruction.
Although~\cite{jeon2023context} obtains competitive results with standard codecs such as VVC~\cite{vvc}, it still relies on complex context modules and iterative RD-prioritized transmission; this makes~\cite{jeon2023context} extremely complex in terms of encoding/decoding time, computational resources, and the number of parameters required.

\section{Proposed method}\label{proposedmethod}

\subsection{General architecture}\label{gen_arch}

\begin{figure}[t!]
  \centering
 \includegraphics[width=0.95\columnwidth]{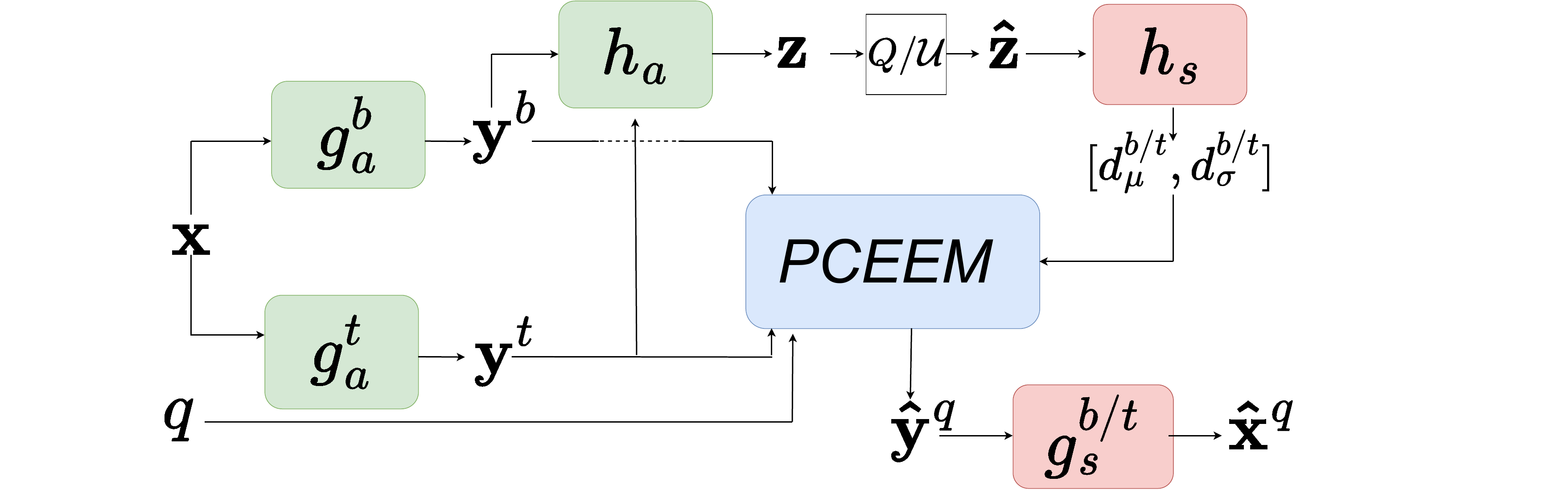}
  \caption{Overview of our proposed architecture. Green and red boxes represent encoder and decoder modules, respectively, while blue boxes must be stored at both encoder and decoder. $q$ represents the target quality.  }
  \label{fig:gen_arch}
\end{figure}

Our progressive method is based on~\cite{zou2022devil} as illustrated in Fig.~\ref{fig:gen_arch}.
It uses a hyperprior-based architecture~\cite{balle2018variational} and a channel-wise entropy model~\cite{minnen2020channel}.
We generate two latent representations: the \emph{base}  provides the reconstructed image at the lowest bitrate, while the \emph{top} yields the image at any higher bitrates; to obtain FGS we first obtain a residual representation by means of element-wise difference between the top and the base latents, and then we introduce a masking system that ranks its elements from the most to least important, and mask only the ones needed to achieve a specific target quality, replacing the others with the mean.
Quality is indicated by $ q\!\in\![0,100]$, where 0 and 100 represent the lowest and highest level, respectively. 

Our approach has two advantages: It makes the first training step easier, since it is sufficient to introduce two different target qualities, i.e. \emph{base} and \emph{top}, in the loss function, and it enables choosing the distance between the lowest and highest levels, thereby focusing on a specific bit range of interest. Furthermore, the masking policy introduced in Sect.~\ref{mask} is simple and efficient, with no parameters to train. 

For an image $\mathbf{x}$ (Fig.~\ref{fig:gen_arch}) the two latent representations $\mathbf{y}^{b/t}$
are extracted with the encoders $g_{a}^{b/t}$, that is, $~{\mathbf{y}^{b/t}\!=\!g_{a}^{b/t}(\mathbf{x};\boldsymbol{\phi}^{b/t})}$, where $\boldsymbol{\phi}^{b/t}$  are learnable parameters, and the superscript $b/t$ represents both the \emph{base} and the \emph{top} representations in a more compact manner.
The latents $\mathbf{y}^{b/t}$ are modeled as Gaussian distributed with mean $\mu^{b/t}$ and standard deviation $\sigma^{b/t}$. As in~\cite{balle2018variational}, a hyperprior autoencoder $~{\!\{h_{a}, h_{s}\}\!}$ is introduced to find spatial correlations within the image, optimizing rate reduction. 
In particular, we have $~{\mathbf{z} = h_{a}(\mathbf{y}^{b}, \mathbf{y}^{t};\boldsymbol{\phi}_{h})}$, which is then quantized to $~{\mathbf{\hat{z}} = Q(\mathbf{z})}$.
$Q$ is the quantization function, which is replaced by the addition of uniform noise ($\mathcal{U}$) during training.
We then obtain $~{[d_{\mu}^{b/t},d_{\sigma}^{b/t}] = h_{s}(\mathbf{\hat{z}},\boldsymbol{\theta}_{h})}$; these tensors are part of the input for the progressive channel-wise entropy estimation module ($\PCEEM$), explained in Sec.~\ref{pweer}, to extract both entropy parameters and  $\mathbf{\hat{y}}^{q}$:
\begin{equation}
    \mathbf{\hat{y}}^{q} = \PCEEM(\mathbf{y}^{b/t},q,[d_{\mu}^{b/t},d_{\sigma}^{b/t}]).
\end{equation}
\noindent
We input $q$ to obtain the latent representation for the target quality $q$, then encode and add them to the final bitstream. 

The final reconstruction $\mathbf{\hat{x}}^q$ at quality $q$ is obtained through the decoders: $~{\mathbf{\hat{x}}^q = g_s^{b/t}(\mathbf{\hat{y}}^{q}, \theta^{b/t})}$, where $\theta^{b/t}$ are learnable parameters. 
We implemented two different decoders that are used depending on the target quality; If $q$ = 0, we use $g_s^{b}$, while in any other case we use $g_s^{t}$.

\begin{figure*}[!t]
  \centering
 \includegraphics[width=0.8\textwidth]{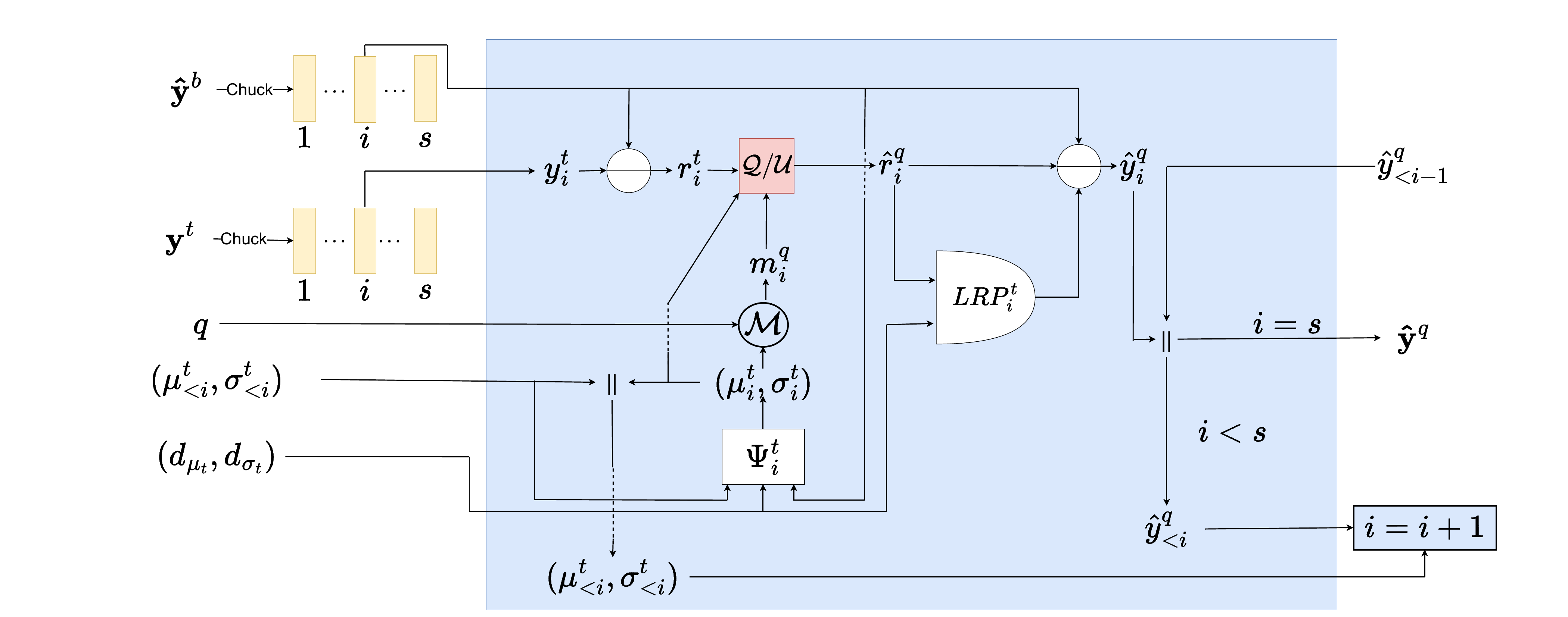}
  \caption{Progressive channel wise entropy estimation model ($\PCEEM$) during the $i$-th slice, considering a general quality $q$. $\mathbf{\hat{y}}^{b}$ represents the base latent representation already obtained.  $||$ represents concatenation along channels, while $\bigodot$ represents element-wise operation, which can be summation (+) or subtraction (-).}
  \label{fig:pceem}
\end{figure*}

\subsection{Progressive channel-wise entropy estimation}\label{pweer}
To develop a \emph{progressive channel-wise entropy estimation model} ($\PCEEM$), both $\mathbf{y}^{b}$ and $\mathbf{y}^{t}$ are divided into $s$ slices along the channels~\cite{minnen2020channel}, for a total of $2\!\times\!s$ slices.
Fig.~\ref{fig:pceem} illustrates $\PCEEM$. 
To achieve the initial base quality reconstruction, only $\mathbf{y}^{b}$ is necessary.
For the other levels, we also consider $\mathbf{y}^{t}$.

To encode the \emph{base} latent, the process is similar to previous works~\cite{minnen2020channel,zou2022devil,liu2023tcm}. Given an index $i$ such that $0 \leq i<s$ and the corresponding slice $y_{i}^{b}$, we obtain the entropy parameters $~{(\mu_{i}^{b},\sigma_{i}^{b}) = \Psi_{i}^{b} (\mu_{i}^{b},\sigma_{i}^{b},\hat{y}_{<i}^{b})}$, where $\Psi_{i}^{b}$ is a learnable module and $\hat{y}_{<i}^{b}$ are the previous slices. 
After the quantization step, $~{\hat{y}_{i}^{b} = Q(y_{i}^{b})}$, a module $\LRP_{i}^{b}$ is applied to reduce the quantization error; the resulting slice $\hat{y}_{i}^{b}$ is concatenated to the previous slices.
The final tensor $\mathbf{\hat{y}}^{b}$ can be used either as input to the decoder $g_{s}^{b}$ to obtain the base reconstruction image $\mathbf{\hat{x}}^{b}$, or as input for estimating the entropy of higher qualities.

For any other possible quality $q$ greater than 0, the \emph{top} latent representation is handled differently.
Given the index $i$ and a target quality $q$, we have the initial slice $y_{i}^{t}$; as in residual video coding~\cite{liu2020deep}, we first compute the latent residual representation $~{r_{i}^{t}\!=\!y_{i}^{t} - \hat{y}_{i}^{b}}$, and then extract the entropy parameters $(\mu_{i}^{t},\sigma_{i}^{t})$ through a learnable module $\Psi_{i}^{t}$: 
\begin{equation} \label{pwerr_entropy_pars}
            (\mu_{i}^{t},\sigma_{i}^{t}) = \Psi_{i}^{t}(\hat{y}_i^{b},d_{\mu}^{t},\mu_{<i}^{t},d_{\sigma}^{t},\sigma_{<i}^{t}).
\end{equation}

Note that $\Psi_{i}^{t}$ is the same for every possible target quality.
Unlike the \emph{base} scenario, this module receives the entropy parameters $(\mu_{<i}^{t}, \sigma_{<i}^{t})$ of previous slices instead of the slices $y_{<i}^{t}$, crucial for applying the masking policy described in Sect.~\ref{mask}, enabling progressive quality representation without altering the already obtained bitstream entropy parameters.
$\sigma_{i}^{t}$ is used to compute the mask $~{m_i^{q}\!=\!\mathcal{M}(\sigma_{i}^{t},q)}$, where $\mathcal{M}$ represents the masking operator.

In summary, $m_i^q$ masks certain elements of $r_i^{t}$, replacing them with the mean $\mu_i^t$.  In this way, 
$r_i^t$ is effectively downsampled, with only one encoded part and the rest replaced with reasonable values such as the mean; this process is described by the following equation:
 
\begin{equation}\label{quant}
    \hat{r}_{i}^{q} = Q(r_i^{t} - \mu_i^t) \otimes  m_i^q + \mu_i^t,
\end{equation} 
where $\otimes$ represents element-wise multiplication.
We first mask $Q(r_i^t - \mu_i^t)$, which is then encoded and added to the bitstream; 
At the decoder side,  we then add back the mean $\mu_i^{t}$ (eq. \ref{quant} );  in this way, all the elements omitted by $m_i^{q}$ are  replaced by the mean already computed. This operation is represented by the red box in Fig.~\ref{fig:pceem}. We employed ANS coding \cite{duda2013asymmetric} for entropy coding; since the residual representation elements are modeled as independent Gaussians with parameters derived from $\Psi_{i}^{t}$, the bitstream can be split into complementary parts without loss of efficiency.

A network $\LRP_{i}^{t}$ is applied to $\hat{r}_{i}^{q}$ to reduce the quantization error~\cite{minnen2020channel}.
Finally, the resulting $\hat{r}_{i}^{q}$ is added to the corresponding base slice $\hat{y}_{i}^{b}$, obtaining the final quantized slice $\hat{y}_{i}^{q}$; the latter is further concatenated (symbol $||$ in Fig.~\ref{fig:pceem}) with the previous slices, obtaining $\hat{y}_{<i}^{q}$. If $i<s$, then the next slice $i+1$ is encoded (blue box $i=i+1$ in Fig.~\ref{fig:pceem}).  

Once all slices have been computed, the resulting tensor $\mathbf{\hat{y}}^{q}$ is used as input for $g_s^{t}$ to obtain the reconstructed image $\mathbf{\hat{x}}^{q}$ at a specific target quality $q$. 

\subsection{Progressive mask-based coding} \label{mask}

To encode elements of $r_{i}^{t}$, we propose a progressive masking method based on the corresponding standard deviation $\sigma_{i}^{t}$; it is worth pointing out that it is possible to compute $\sigma_{i}^{t}$ on both the encoder and decoder sides, as they are required for the encoding / decoding of the bitstream, and also it does not change with respect to different qualities.  
The general idea is to add more and more elements of $r_{i}^{t}$ to progressively improve reconstruction quality.
In this context, it is reasonable to identify the most important elements in the residual latent representation as those with a larger variance, as they are more likely to change and cause a greater reconstruction error; this is the reason we use $\sigma_{i}^{t}$.

In particular, a progressive masking procedure for the $i$-th  block $r_{i}^{t}$ is described in Alg.~\ref{alg}.
Each target quality value $~{q\!\in\![0,100]}$ unambiguously determines the elements of $r_{i}^{t}$ to be masked, i.e. skipped, in the encoded bitstream. A higher $q$ means fewer masked elements.
We compute the $~{(100 - q)}$-th percentile of $\sigma_{i}^{t}$ and mask all elements of $r_{i}^{t}$ whose corresponding standard deviation is below that value (lines 7-12 in Alg.~\ref{alg}). 
These elements are not encoded in the bitstream, but are replaced with the mean $\mu_{i}^{t}$ (line 16).

This approach first encodes elements with higher standard deviation, improving reconstruction quality. The latent representation is divided into complementary portions, prioritized by importance, which can be decoded separately.
This masking policy has multiple advantages:
\begin{itemize}[itemsep=0.05em] 

    \item It is an easy-to-compute progressive mechanism to obtain RD transmission priority, 
    since it does not require any optimization procedure or further parameters; 
    \item It is possible to compute such masks consistently both at the encoder and the decoder, providing also elements position: this allows the encoder to simply skip the masked values without any signaling overhead. 
    \item Only the value $q$ needs to be signaled within the encoded bitstream. However, $q$ is represented by a floating point or an integer with negligible overhead.
\end{itemize}


\begin{algorithm}
\caption{$\boldsymbol{\sigma}$-masking procedure. $\mathcal{M}$}\label{alg}
\begin{algorithmic}[1]

\INPUT $\sigma_{i}^{t}$, \hspace{0.1cm} $\mu_{i}^{t}$, \hspace{0.1cm} $r_{i}^{t}$, \hspace{0.1cm}  $q$
\OUTPUT $m_{i}^{q}$ \hspace{0.1cm} $\hat{r}_{i}^{q}$

\State  $sh \gets Shapes(\sigma_{i}^{t})$  \Comment{extract the shape of $\sigma_{i}^{t}$}
\State $\sigma_{i}^{t}  \gets Flatten(\sigma_{i}^{t})$ \Comment{flatten $\sigma_{i}^{t}$ to a 1-dim vector }
\State $L \gets len(\sigma_{i}^{t})$ \Comment{extract length of $\sigma_{i}^{t}$}
\State $q$-\text{th percentile}  $\gets \mathrm{Perc}(\sigma_{i}^{t}, 100 - q)$ 
\State $m_{i}^{q} \gets Zeros(L)$
\State  $j \gets 0$
\While{ $j < L$}
\If{$\sigma_{i}^{t} < q$-th percentile}
    \State  $m_{i}^{q}[j] = 0$
\Else{}
     \State $m_{i}^{q}[j] = 1$
\EndIf
\State $j \gets j + 1$
\EndWhile
\State $m_{i}^{q} \gets Reshape(m_{i}^{q}, sh)$ \Comment{reshape $m_{i}^{q}$ to \emph{sh}}
\State $\hat{r}_{i}^{q} \gets  Q(r_{i}^{q} - \mu_{i}^{t})*m_{i}^{q} + \mu_{i}^{t}$ 
\end{algorithmic}
\end{algorithm}

\begin{figure}[!h]
  \centering
 \includegraphics[width=1\columnwidth]{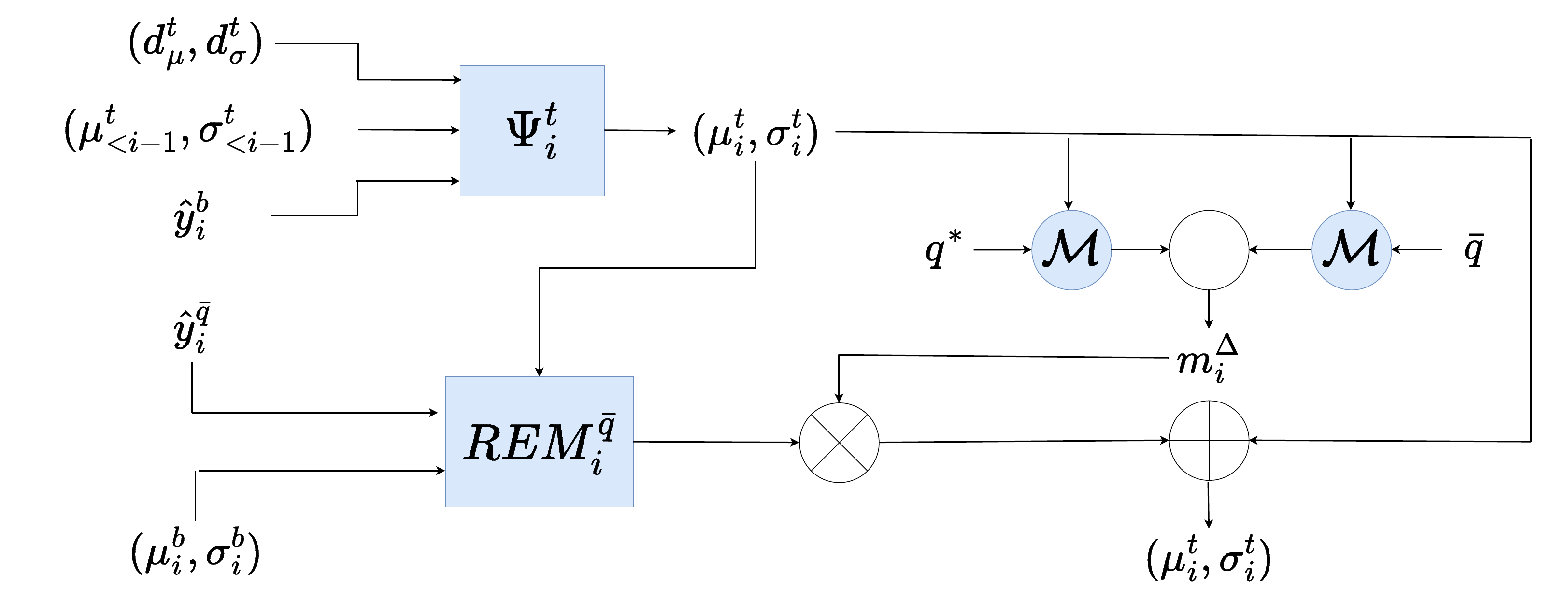}
  \caption{Blueprint of REM for a fixed checkpoint quality $\bar{q}$ and for the slice $i$.}
  \label{fig:remblueprint}
\end{figure}

\subsection{Slice-wise rate enhancement modules} \label{rem}
Taking inspiration from~\cite{jeon2023context}, we introduce checkpoint-based learnable modules by means of \emph{Slice-wise Rate Enhancement Modules} (REMs) that aim to improve the estimation of the entropy parameters of each slice.
The idea is to fix some checkpoint qualities along the bit range and use them as input for a REM to improve the encoding of higher quality.
As explained in Sect. \ref{pweer}, regarding the \emph{top} representation, we do not utilize slices $\hat{y}_{i}^{t}$ as in \cite{minnen2020channel}; REMs are meant to incorporate information coming from already encoded/decoded residual representation.

Consider a specific slice $i$, a checkpoint quality $\bar{q}$, and a target quality $q^*$, with $q^* > \bar{q}$. Fig.~\ref{fig:remblueprint} shows how we inserted such a module, corresponding to the $i$-th slice and checkpoint quality $\bar{q}$, which we called $REM_{i}^{\bar{q}}$.
It receives three elements as input: the quantized representation of the checkpoint $\mathbf{\hat{y}_{i}^{\bar{q}}}$ (already decoded and present in memory), and $(\mu_{i}^{k},\sigma_{i}^{k})$ with $k\!=\!\{b,t\}$, and enhances the estimation of $(\mu_{i}^{t}, \sigma_{i}^{t})$ only for residual latent elements that must be added to transition from checkpoint quality $\bar{q}$ to $q^*$.
This is represented by the mask $m_i^{\Delta}$ in Fig.~\ref{fig:remblueprint}, computed as:

\begin{equation}
    m_{i}^{\Delta} = \mathcal{M}(\sigma_{i}^{t}, q^*) - \mathcal{M}(\sigma_{i}^{t}, \bar{q}).
\end{equation}
which preserves only those points that need to be added progressively for the quality $q^*$; this means that the $REM_{i}^{\bar{q}}$  will affect only the elements that have been added to the bitstream, leaving unchanged the parameters of the other elements. 
In practice, by fixing some checkpoint qualities  $\bar{q}$'s for improving rate estimation, we divide the bit range into different subranges, denoted by the symbol  $L_b$ if only $\mathbf{\hat{y}}^{b}$ is used (\emph{base} level), $L_0$ if no REMs are used, and $L_{\bar{q}_{j}}$ if the $i$-th REM$^{\bar{q}_{j}}$ is used; here $j$ travels across the number of checkpoints added.

\begin{figure*}[!t]
  \centering
 \includegraphics[width=1\textwidth]{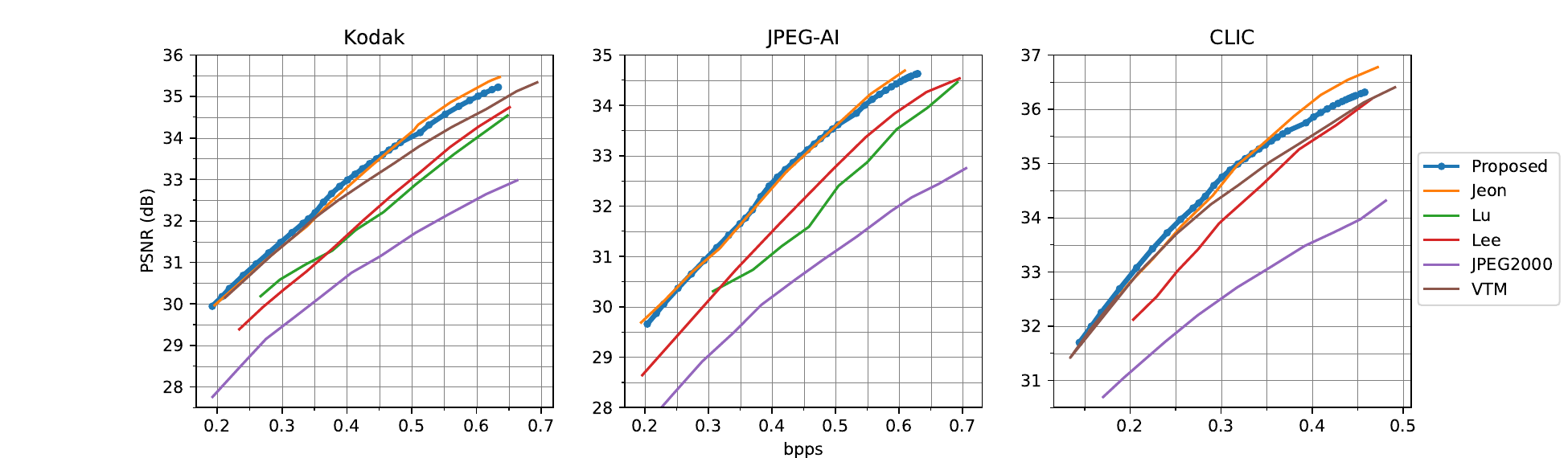}
  \caption{Rate-distortion performance of our method compared with progressive image compression algorithms: Jeon \cite{jeon2023context}, Lee \cite{lee2022dpict}, Lu \cite{lu2021progressive} and JPEG2000 \cite{jpeg2k}. We tested our method on Kodak (left), JPEG-AI (center), and CLIC validation dataset (right).}
  \label{fig:exp}
\end{figure*}

\subsection{Training}\label{training}
Training occurs in multiple phases.
The first phase is to train the general architecture without any REMs to reconstruct the image at the base and top qualities. The loss function is the sum of the \emph{rate-distortion} trade-off for the two different levels:
\begin{equation} \label{eq_loss}
\mathcal{L} =  \sum_{k = \{b,t\}}\mathrm{L}_{k},
\end{equation}
where each term of the summation is represented by
\begin{equation} \label{loss:each}
\begin{split} 
    \mathrm{L}_{k} &= \lambda_{k}\mathbb{E}_{\mathbf{x}\sim p_{\mathbf{x}}}[d(\mathbf{x}, \mathbf{\hat{x}}^{k})]  + \\ 
         &\mathbb{E}_{\mathbf{x}\sim p_{\mathbf{x}}}[-\log_{2} p_{\mathbf{\hat{y}}^{k}|\mathbf{\hat{z}}}({\mathbf{\hat{y}}^{k}|\mathbf{\hat{z}}}) - \log_{2}p_{\mathbf{\hat{z}}} (\mathbf{\hat{z}})].
\end{split}
\end{equation}

The first term in \eqref{loss:each} represents the quality of the reconstruction, where $d$ represents a generic distortion metric, while the other two terms estimate the bitrate required for the two latent representations $\mathbf{\hat{y}}^k$ and $\mathbf{\hat{z}}$.
Here, $\boldsymbol{\lambda}\!=\!\{\lambda_{b},\lambda_{t}\}$ denotes the Lagrangian parameters that balance the trade-off between rate and quality.

The second phase refines $g_{s}^{t}$ to be able to reconstruct images with different qualities $0 < q\leq 100$. When $q$=100 we consider the \emph{top} latent representation without the mask. In this context, we only optimize the first term of \eqref{loss:each}, considering only $k\!=\!t$.
In particular, in this stage, at each training step, we sample a value of $q$ and refine $g_{s}^{t}$ for that specific target quality; this enhances the decoder's robustness to various values of $q$, representing different versions of the residual latent representation. 

Finally, we insert a fixed number of REMs at certain checkpoint qualities and train them in the architecture, keeping other network parts frozen. In this phase, the training loss is the second term of \eqref{loss:each} considering $k\!=\!t$. The number of REM modules is arbitrary, trading off the complexity of the entire architecture and the rate reduction.

\section{Experiments}

\begin{figure*}[!ht]
  \centering
 \includegraphics[width=0.7\textwidth]{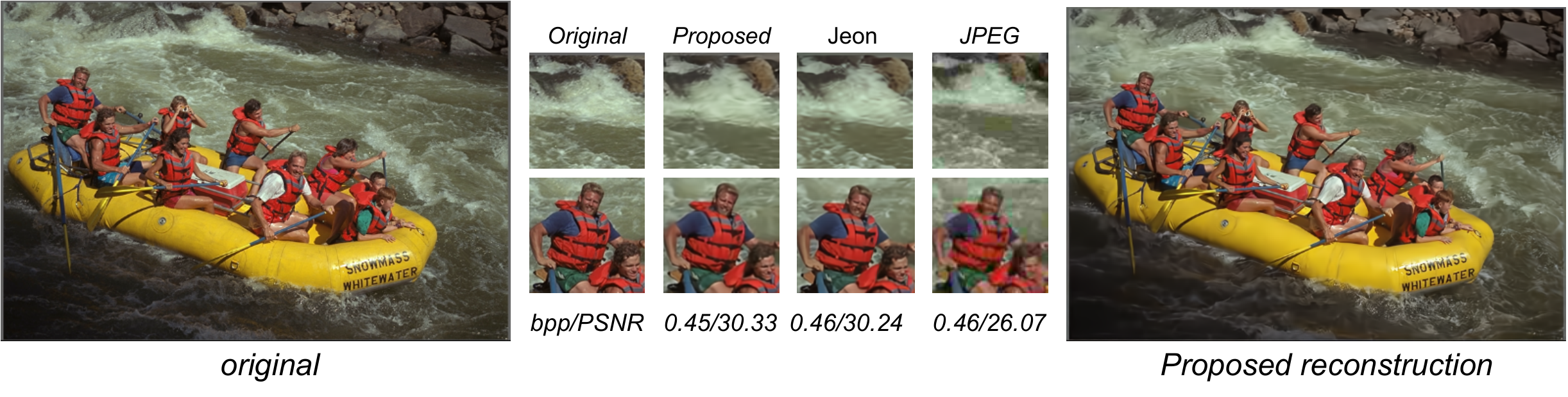}
  \caption{Kodim14 reconstruction from Kodak dataset by different codecs: Proposed, Jeon~\emph{et~al.}, JPEG.}
  \label{fig:sample}
\end{figure*}

\subsection{Setup and training details}
\label{setup}

For training, we used 300k images from the OpenImages dataset~\cite{openimages} and the Adam optimizer, fixing the dimensions of $\mathbf{y}^{b/t}$ and $\mathbf{z}$ to 320 and 192, respectively.
In the first training phase, we trained the entire model for 100 epochs with a batch size of 16 and a learning rate of $10^{-4}$, which was reduced by a factor of two after a plateau with patience of ten epochs. We set $\boldsymbol{\lambda} = \{5 \times 10^{-3}, 5 \times 10^{-2}\}$ in \eqref{loss:each}.

Secondly, we refine $g_s^{t}$ for 80 epochs with a learning rate of $10^{-4}$, while in the last one we fixed the number of REMs to three to balance performance and complexity, empirically selecting quality checkpoints at $~{\bar{q}\!=\!\{0.5, 7.5, 20\}}$ and refining them for 30 epochs.
These values, approximately 5\%, 40\%, and 65\% of the Kodak bitstream, ensure that a REM module covers the entire bit range while improving entropy parameter estimation in a sensible portion of the latent space, avoiding excessively high $q$.

We compare our method with several models (Jeon~\emph{et~al.}~\cite{jeon2023context}, Lee~\emph{et~al.}~\cite{lee2022dpict}, Lu~\emph{et~al.}~\cite{lu2021progressive}, and JPEG2000~\cite{jpeg2k}), on three datasets: Kodak~\cite{kodak}, CLIC validation dataset~\cite{clic}, and JPEG-AI~\cite{jpeg_ai}. 
Our implementation is built upon the \texttt{compressai} \cite{compressai} library, and we used  NVIDIA A40 gpu for training.\footnote{Source code is available at \href{https://github.com/EIDOSLAB/Efficient-PIC-with-Variance-Aware-Masking}{https://github.com/EIDOSLAB/PIC}}

\subsection{Rate-distortion (RD) performance}\label{rd}
Fig.~\ref{fig:exp} shows the PSNR results for the three datasets.
Our method outperformed three of the other algorithms in the bit range and is competitive with Jeon~\emph{et~al.}, as seen in the reconstruction in Fig.~\ref{fig:sample}.
In particular, we outperformed Jeon~\emph{et~al.} for low and medium bit rates, with a slight deterioration for higher qualities. 
The competitiveness of our model is also seen in Table~\ref{tab_bdrate}, which represents BD-Rate and BD-PSNR for the three different datasets considering Jeon~\emph{et~al.} as the reference model. Our method is better for Kodak and CLIC, with a small deterioration for JPEG-AI. 

\begin{figure}[!ht]
  \centering
 \includegraphics[width=0.8\columnwidth]{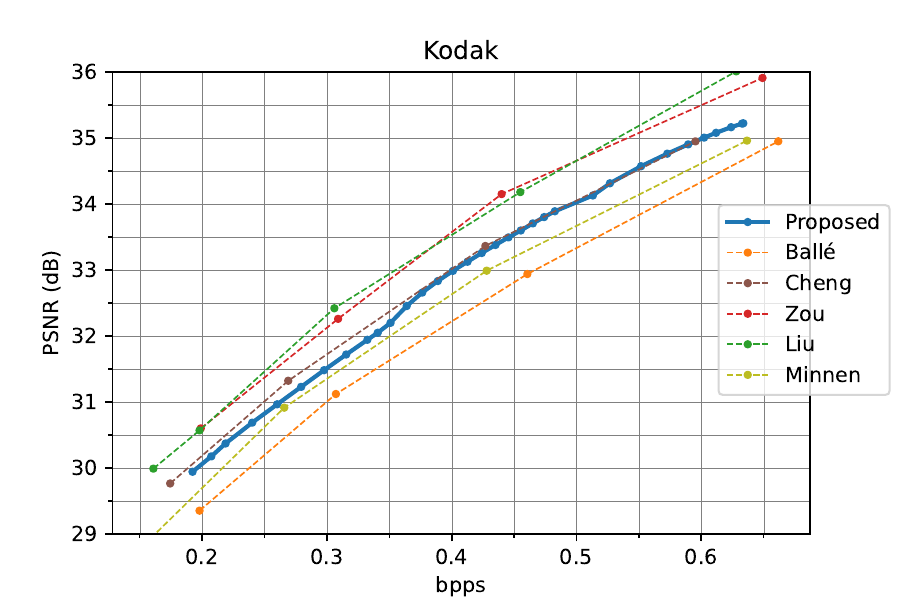} \vspace{-0.2cm}
  \caption{RD curve for some fixed-rate models on Kodak: Ballé~\emph{et~al.}\cite{balle2018variational}, Minnen~\emph{et~al.}\cite{minnen2018joint} Cheng~\emph{et~al.}\cite{cheng2020learned},  Zou~\emph{et~al.}\cite{zou2022devil}, and Liu ~\emph{et~al.}\cite{liu2023tcm}. Dotted lines represent that such models are not able to achieve FGS. }
  \label{fig:rd_fixed}
\end{figure}

Fig.~\ref{fig:rd_fixed} shows the RD trade-off with respect to some state-of-the-art fixed-rate (non-progressive) models. As these methods do not have the progressive property, our architecture is expected not to outperform them.
Although we performed less well than the most recent models (for example, Zou~\emph{et~al.}~\cite{zou2022devil} and Liu~\emph{et~al.}~\cite{liu2023tcm}), we obtained comparable results with other methods,  demonstrating our competitiveness. We point out that the deep learning models in Fig.~\ref{fig:rd_fixed}  are fixed-rate, i.e., there is a separate model for each target quality, and they do not achieve FGS.

\begin{table}
\centering
\small
 \caption{BD-Rate and BD-PSNR  on the three datasets considering Jeon~\emph{et~al.}  as reference and our method as proposed.   }
 \vspace{-0.2cm}
\resizebox{0.6\columnwidth}{!}{
\begin{tabular}{p{0.16\linewidth}ccc} 
  \toprule
   & \textbf{\emph{Kodak}}& \textbf{\emph{JPEG-AI}}& \textbf{\emph{CLIC}}   \\
  \midrule
  \emph{BD-Rate}& $\mathbf{-1.05}$ & 0.47 & $\mathbf{-0.75}$  \\
  \emph{BD-PSNR}& $\mathbf{0.04}$ &  -0.01 & $\mathbf{0.01}$  \\

  \bottomrule
\end{tabular}
}
\label{tab_bdrate}
\end{table}

\begin{figure}[t]
    \centering
    \begin{minipage}[b]{0.30\textwidth}
        \centering
        \includegraphics[width=0.95\textwidth]{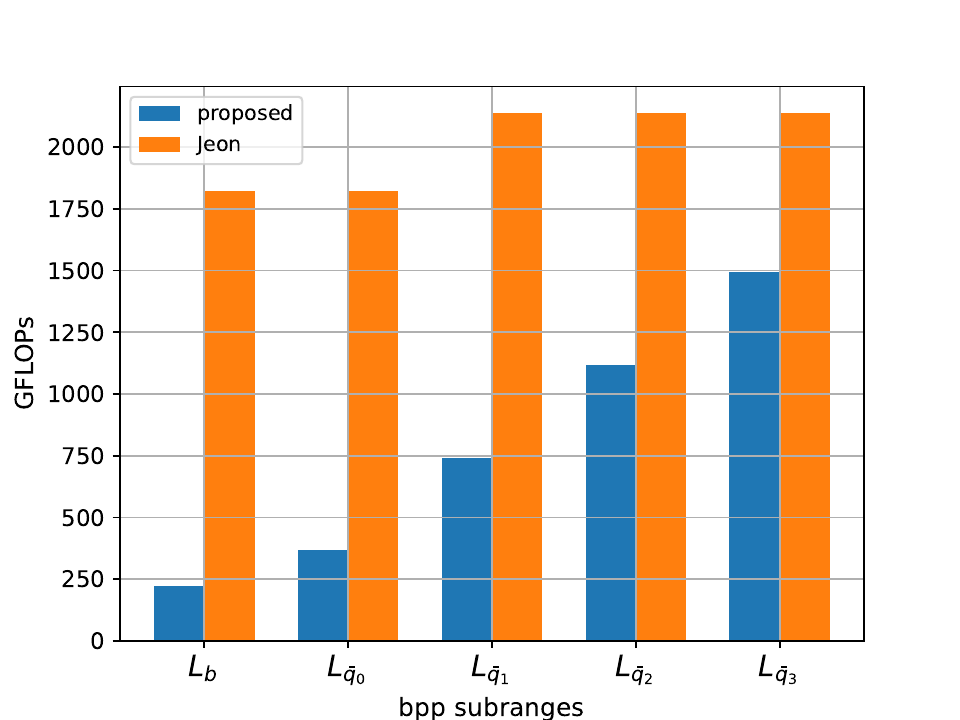}
        \subcaption{ }
        \label{fig:com1}
    \end{minipage}
    \vspace{-0.25cm}
    \begin{minipage}[b]{0.30\textwidth}
        \centering
        \includegraphics[width=0.95\textwidth]{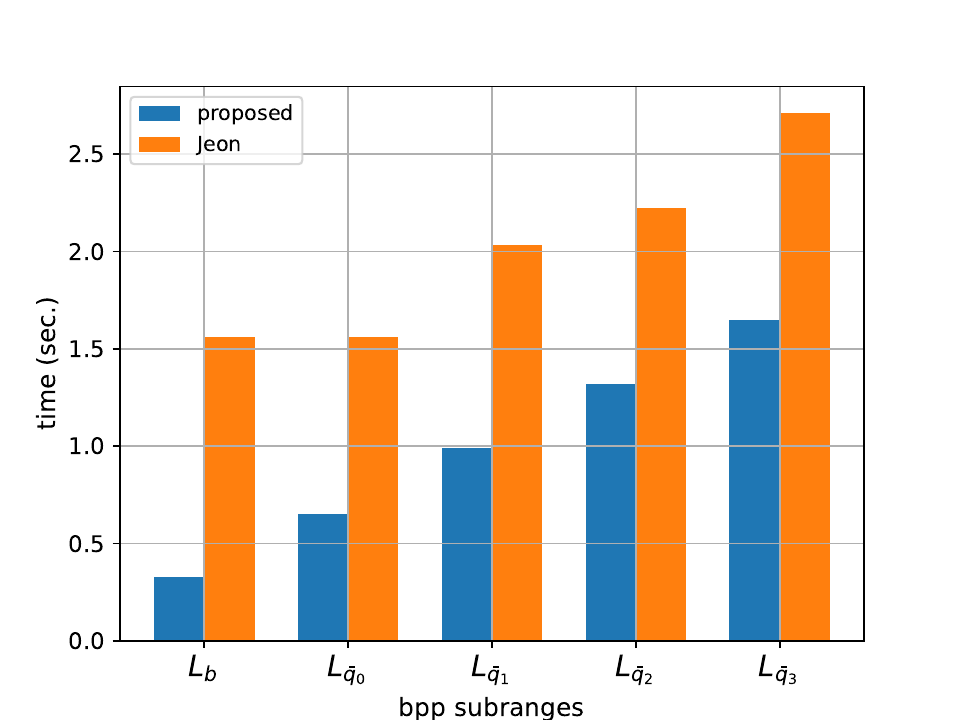}
        \subcaption{}
        \label{fig:com2}
    \end{minipage}
    \caption{GFLOPs and decoding time complexity on NVIDIA A40 GPU (a,b)  vs.\ Jeon~\emph{et~al.} for different subranges on Kodak.}
    \label{com}
\end{figure}

\begin{table*}[!h]
\caption{Complexity comparison between our proposed method and Jeon~\emph{et~al.} on Kodak.} 
\vspace{-0.3cm}
\centering
\resizebox{0.75\textwidth}{!}{
\begin{tabular}{l   l l l  | l l l  | l l   }
\toprule
 \multicolumn{1}{c}{}  & \multicolumn{3}{c|}{\textbf{\emph{GPU}}}  & \multicolumn{3}{c|}{\textbf{\emph{CPU}}} & \multicolumn{2}{c}{\textbf{\emph{Parameters} (M)}} \\
\cmidrule(l){2-4} \cmidrule(l){5-7} \cmidrule(l){8-9}
 & Dec. time  &   Enc. time   & 
$\sim$ GFLOPs   &  Dec. time      &  Enc. time  & 
GFLOPs &  Encoder  &   Decoder    \\
\midrule

\emph{Proposed}  & 1.15  & 1.72  & 787.94 &  6.33  & 5.49  &789.98 &  97.26  & 90.78  \\
Jeon~\emph{et~al.}  &  2.018  & 1.64   &2012 & 10.9  &  4.16    &1980.87 &  323.6   & 399  \\

\bottomrule
\end{tabular} \label{tab:complexity}
}
\end{table*}

\subsection{Model Complexity} \label{complexity_sec}
Figures \ref{fig:com1} and \ref{fig:com2} represent GFLOPs and the time to decode on GPU, respectively; we compared our results with the most complex version of Jeon~\emph{et~al.}, which is competitive in terms of RD.
 
We analyze the different ranges defined in Sect.~\ref{rem}, to understand how complexity varies when adding additional portions of the bitstream.
With our efficient masking system and parallelizable modules, our method is optimal in time and computational resources compared to Jeon~\emph{et~al.}.

The complexity improvement of our method is substantial, especially for lower bitrates, i.e., where fewer modules, and so fewer iterations, are required. Our gain marginally decreases when increasing the bit rate while always being significantly faster than Jeon~\emph{et~al.}. On average, we obtain an improvement of a factor of $2$ both in terms of GFLOPs and decoding time, as shown in Figures ~\ref{com}.

Our method becomes less efficient at higher qualities due to REMs, as decoding the fixed checkpoint latent representations is necessary before achieving the target quality.
We attribute our superior time performance compared to Jeon~\emph{et~al.} to several factors: first, our general architecture, with also the addition of REMs, is smaller than the context model used by Jeon~\emph{et~al.} and only needs to be applied once during decoding; 
furthermore, our masking policy is fast and efficient, whereas Jeon~\emph{et~al.} relies on a complex optimization method for RD priority transmission; 
finally, our algorithm allows one to encode and decode the complementary portions of the final bitstream, making the computation simpler for a specific target quality.
Tab.~\ref{tab:complexity} compares the complexity of our method and Jeon~\emph{et~al.} on the entire bit range of interest. Our method is better for both GPU and CPU usage,  requires fewer parameters, and maintains similar performance at encoding.

\begin{figure}[ht!]
  \centering
 \includegraphics[width=0.8\columnwidth]{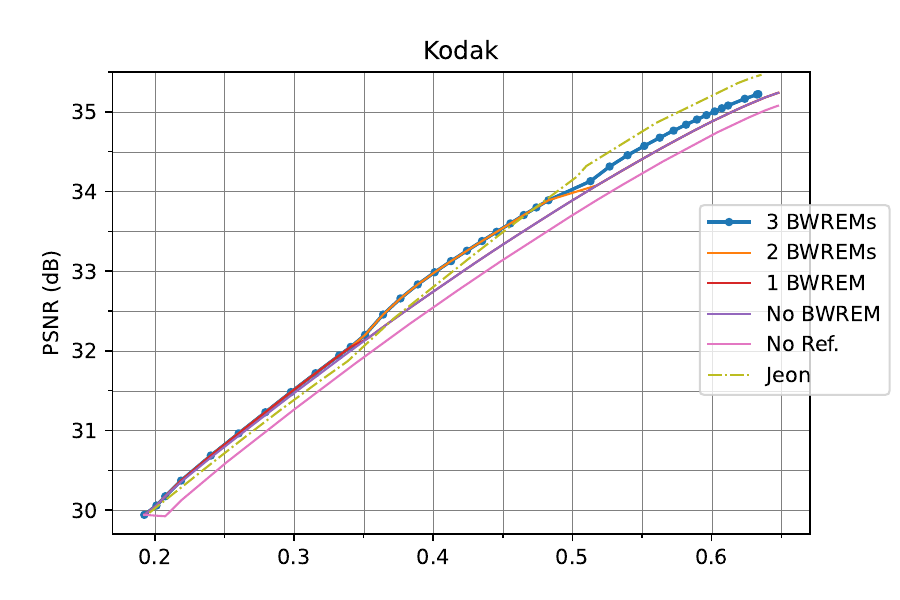}
\vspace{-0.4cm}
  \caption{Results on Kodak considering different configurations. } 
  \label{fig:rd_ablation}
\end{figure}

\begin{table}
\centering
\small
 \caption{BD-RATE, BD-PSNR, and the number of parameters for different configurations considering Jeon~\emph{et~al.} as the base model.}
 \vspace{-0.3cm}
\resizebox{0.93\columnwidth}{!}{\begin{tabular}{p{0.35\linewidth}ccc} 
  \toprule
   & \textbf{\emph{BD-Rate}}& \textbf{\emph{BD-PSNR}}&  \textbf{ \# \emph{ Dec. Pars (M)}}   \\
  \midrule
  \emph{3 REMs + $g_s^{t}$ Refine}& $\mathbf{-1.05}$ & $\mathbf{0.04}$ &  90.8   \\
   \emph{2 REMs + $g_s^{t}$ Refine}& $\mathbf{-0.56}$ & $\mathbf{0.02}$ &  79.4  \\
    \emph{1 REM + $g_s^{t}$ Refine }& 0.94 & - 0.04 &   67.9   \\
     \emph{ $g_s^{t}$ Refine}& 1.24 & - 0.05 &  56.6   \\
      \emph{No $g_s^{t}$ Refine}& 5.53 & -0.25 &  56.6   \\
  \bottomrule
\end{tabular}}\label{tab_ablation}
\end{table}

\subsection{Ablation study} \label{ablation}

Fig.~\ref{fig:rd_ablation} shows how REMs and decoder retraining improve the performance of our method on Kodak.
The refinement of $g_{s}^{t}$ turns out to be important in improving overall performance (purple line versus pink line in Fig.~\ref{fig:rd_ablation}); this is reasonable since we make the decoder robust to change of qualities.
Furthermore, adding more REMs enhances coding efficiency across various subranges of the bitrate.
The module operating in the central portion of the bit range offers the most significant improvement because at this stage in progressive coding, a large portion of the bitstream is already available, meaning the checkpoint quality has more information compared to lower-quality stages, but there is still room for enhancement in the remaining bitstream.

Table~\ref{tab_ablation} shows results in terms of BD-Rate and BD-PSNR, using Jeon~\emph{et~al.}, as reference model; The proposed method produces a gain in terms of general RD performance only from 2 REMs, always exploiting less parameters.

\begin{figure}[!h]
    \centering
    \begin{minipage}[t]{0.45\textwidth}
        \centering
        \includegraphics[width=0.43\textwidth]{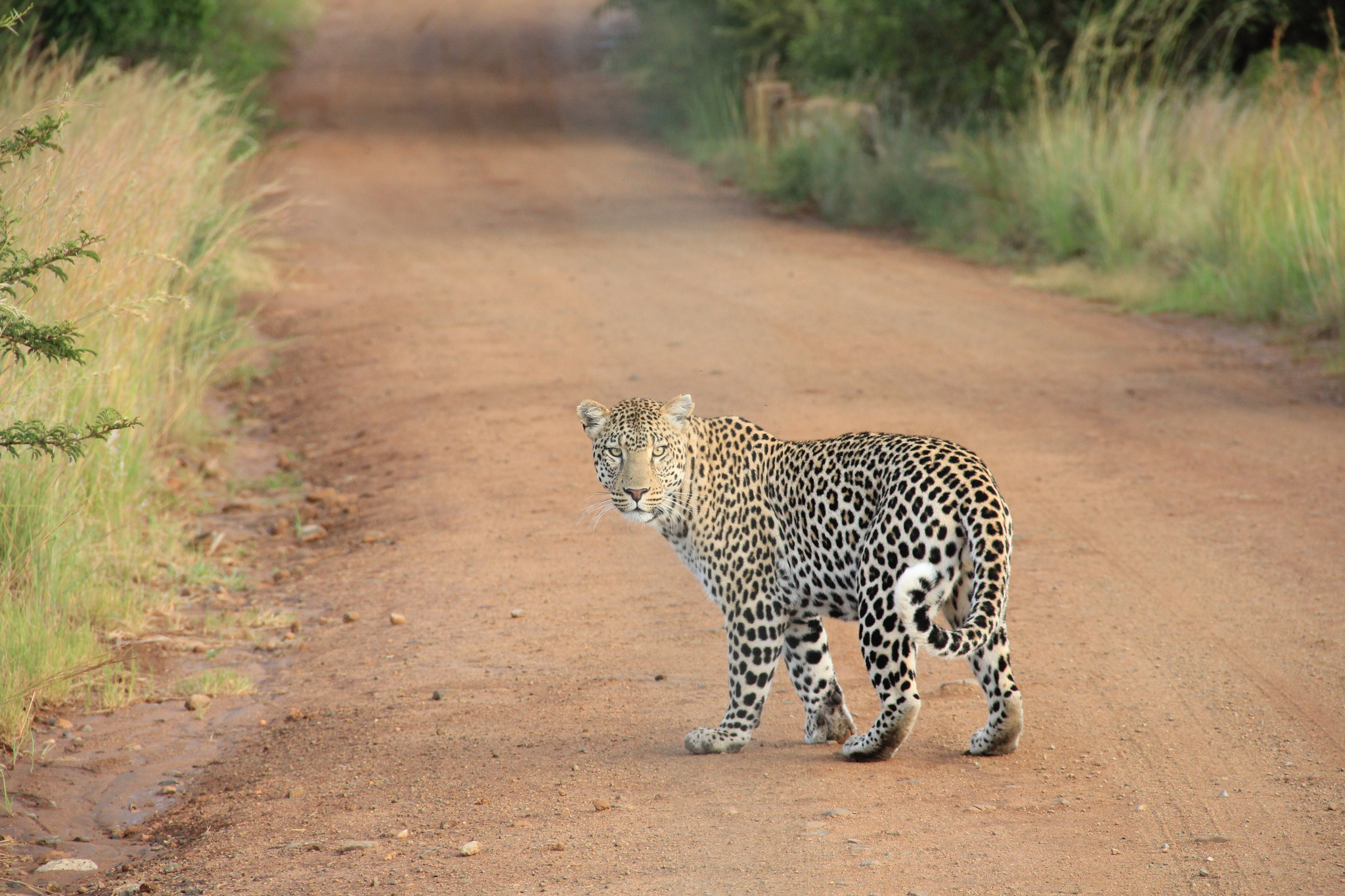}
        \subcaption{Input image}\label{fig:big-image}
        \vspace{0.2cm}
        \begin{minipage}[t]{0.40\textwidth}
            \centering
            \includegraphics[width=0.68\textwidth]{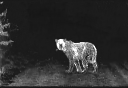}
            \subcaption{Base std}\label{fig:small-image1}
        \end{minipage}
        \hfill
        \begin{minipage}[t]{0.40\textwidth}
            \centering
            \includegraphics[width=0.68\textwidth]{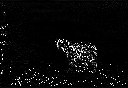}
            \subcaption{Base channel}\label{fig:small-image3}
        \end{minipage}
    \end{minipage}
    \vfill
    \begin{minipage}[t]{0.48\textwidth}
        \centering
        \begin{minipage}[t]{0.26\textwidth}
            \centering
            \includegraphics[width=0.93\textwidth]{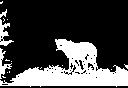}
            \subcaption{mask at $q$=10}\label{fig:image4}
        \end{minipage}
        \hfill
        \begin{minipage}[t]{0.26\textwidth}
            \centering
            \includegraphics[width=0.93\textwidth]{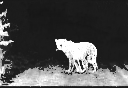}
            \subcaption{std at $q$=10}\label{fig:image5}
        \end{minipage}
        \hfill
        \begin{minipage}[t]{0.26\textwidth}
            \centering
            \includegraphics[width=0.93\textwidth]{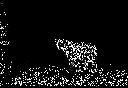}
            \subcaption{channel at $q$=10}\label{fig:image6}
        \end{minipage}
        \vfill
        \begin{minipage}[t]{0.26\textwidth}
            \centering
            \includegraphics[width=0.93\textwidth]{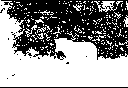}
            \subcaption{Mask at $q$=50}\label{fig:image7}
        \end{minipage}
        \hfill
        \begin{minipage}[t]{0.26\textwidth}
            \centering
            \includegraphics[width=0.93\textwidth]{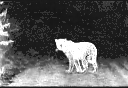}
            \subcaption{std at $q$=50}\label{fig:image8}
        \end{minipage}
        \hfill
        \begin{minipage}[t]{0.26\textwidth}
            \centering
            \includegraphics[width=0.93\textwidth]{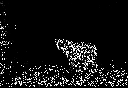}
            \subcaption{channel at $q$=50}\label{fig:image9}
        \end{minipage}
    \end{minipage}
    \vspace{-0.3cm}

    \caption{Original image (a), base latent channel and estimate standard deviation (b,c) obtained mask, estimate standard deviation, and resulting latent channel for two different q (d,e,f,g,h,i). }\label{fig:main}
\end{figure}

\subsection{Masking policy } \label{mask_p}
\vspace{-0.1cm}


Fig.~\ref{fig:main} shows how our masking policy works for different $q$'s, considering a sample image from the CLIC validation dataset. 
As expected, the base level is characterized by very low std values (Fig.~\ref{fig:main}~(b)) for less variable areas, i.e., the background, with the model focusing only on the main part of the image, such as the foreground.
As $q$ increases, more elements are added. Following the policy introduced in Sec.~\ref{mask}, the most relevant elements, which are part of the subject in the foreground, are prioritized (Fig.~\ref{fig:main}~(d)), and then those that have less impact on the final reconstruction (\ref{fig:main}~(g,j)).
In this way we obtain latent representations with increasing details (Fig.~\ref{fig:main}~(f,i,l)), resulting in reconstructions at multiple quality levels.

\section{Conclusions}

We proposed a novel learning-based model for progressive
image compression based on two latent representations: \emph{base} for the lowest quality reconstruction and \emph{top} for all higher qualities.
FGS is achieved by a lightweight masking system that decomposes the residual between the
top and the base latent representations into complementary parts that can be encoded/decoded separately.
We also introduced
REM modules in the proposed architecture, improving
entropy estimation.

Despite simplicity,  our method achieves competitive, if not better,  RD results with SOTA methods Jeon~\emph{et~al.}~\cite{jeon2023context}, Lee~\emph{et~al.}~\cite{lee2022dpict}, reducing computational complexity, decoding time, and required parameters.

Given the promising results, this method could be further improved, for example, by introducing a system for calculating entropy parameters that takes advantage of the lower
quality levels continuously, without the need to set checkpoints
or REMs, or by introducing a scalable system for the
\emph{base} layer, to obtain both granularity on it and a larger bitrange.

\section*{Acknowledgements}
This research was partially funded by Hi!PARIS Center on Data Analytics and Artificial Intelligence.
{\small
\bibliographystyle{ieee_fullname}
\bibliography{main}
}

\end{document}